\documentclass[journal]{IEEEtran}

\usepackage{cite}
\usepackage{amsmath,amssymb,amsfonts}   %
\usepackage{graphicx}
\usepackage{textcomp}
\usepackage{multirow}
\usepackage{url}
\usepackage{float}                       %
\usepackage{booktabs}                    %
\usepackage{makecell}
\usepackage{xcolor}                      %
\newcommand{\Skip}[1]{}

\begin{document}

\title{CORAL: Curriculum-Optimized Reward Adaptation for LiDAR-Based Goal-Directed Urban Driving}

\author{Anisa~Saleem
        and~Duksu~Kim
\thanks{The authors are with the Department of Computer Engineering, Korea University of Technology and Education (KOREATECH) (e-mail: anisasaleem@koreatech.ac.kr; bluekds@koreatech.ac.kr). \emph{(Corresponding author: Duksu Kim.)}}%
%\thanks{This work was supported by the Institute of Information and Communications Technology Planning and Evaluation (IITP) grant funded by the Korea Government (MSIT) (Grant 2019-0-00001).}%
%\thanks{ORCID: A.\ Saleem, 0009-0004-2288-9178; D.\ Kim, 0000-0002-9075-3983.}
}

\markboth{Saleem and Kim: CORAL---Curriculum-Optimized Reward Adaptation for LiDAR-Based Urban Driving}
{Saleem and Kim: CORAL---Curriculum-Optimized Reward Adaptation for LiDAR-Based Urban Driving}

\maketitle

\begin{abstract}
Reinforcement learning is promising for autonomous urban driving, but long-horizon goal-directed navigation asks a policy to acquire several competing behaviors at once---reaching a distant goal, tracking a route, avoiding obstacles, obeying signals---and a fixed objective gives no order in which to learn them. This paper presents CORAL, which advances two schedules together: a five-stage curriculum that progressively lengthens routes and tightens behavioral constraints, and a stage-aware reward whose component weights shift emphasis from mission progress toward route following, safety, smoothness, and rule compliance as the task hardens. The policy is a multi-stream actor--critic network trained with Proximal Policy Optimization (PPO) in CARLA on a compact $99$-dimensional state pairing a polar LiDAR histogram with vehicle telemetry, ego-frame route geometry, and traffic-rule indicators---no point-cloud encoder, no bird's-eye-view rasterization. Against two PPO baselines under an identical protocol, CORAL reaches the goal in all twenty evaluation episodes on the longest routes under the full set of behavioral constraints, where the baselines reach $5\%$ and $10\%$; a factorial ablation shows that neither schedule alone matches their combination: removing either lowers both success and route completion, and disabling both drops success to $55\%$. Trained in one town, the policy transfers zero-shot to seven unseen towns, succeeding in $68$--$98\%$ of episodes on routes of the same $100$--$150\,\mathrm{m}$ length, with mean lateral deviation below $0.35\,\mathrm{m}$.
Scheduling the objective in step with task difficulty, on a compact state, is thus an effective basis for goal-directed urban driving without visual input or expert demonstrations.
\end{abstract} 

\section{Introduction}
\IEEEPARstart{A}{utonomous} driving in urban environments demands robust perception, long-horizon planning, and smooth continuous control amid diverse conditions~\cite{delavari2025comprehensive}. Traditional imitation learning (IL) methods rely on expert demonstrations, which constrains generalization to novel scenarios or complex intersections~\cite{luimitation}. Although IL performs well when expert data are abundant, it struggles under distribution shift, rare corner cases, and unseen road geometries where no demonstrations exist, making it insufficient for goal-directed navigation across diverse urban layouts.

Reinforcement learning (RL) offers an alternative, allowing agents to learn driving behaviors through direct interaction in simulators such as CARLA~\cite{dosovitskiy2017carla}. RL is appealing because it avoids the need for expert labels and can optimize long-horizon objectives such as route completion and smooth control. However, model-free RL for realistic driving remains difficult, facing high-dimensional observations, sparse rewards, and training instability~\cite{delavari2025comprehensive}.

A further key design choice in RL-based driving is the sensory modality. Vision-based approaches prevail in the literature~\cite{chen2024end}, but RGB inputs are vulnerable to illumination changes, occlusions, and adverse weather~\cite{li2023emergent,zhang2023perception}, whereas LiDAR provides geometric measurements that are largely invariant to illumination---at the cost of computationally heavy raw point clouds or lossy bird's-eye-view (BEV) rasterizations (Section~\ref{sec:related}). These trade-offs motivate a compact, geometry-based LiDAR representation tailored for stable RL~\cite{cai2021carl}.

To this end, we propose \textbf{CORAL} (Curriculum-Optimized Reward Adaptation for LiDAR-based goal-directed urban driving), whose central element is a training schedule that advances task difficulty and the reward objective together. CORAL runs on a compact route-aware observation (Section~\ref{subsec:obs}): rather than raw point clouds or dense BEV maps, the agent acts on a low-dimensional state that combines a 64-bin polar LiDAR histogram with vehicle telemetry, the upcoming route expressed in the ego frame, and traffic-rule signals, giving an efficient description of the driving scene at a fraction of the input dimension that learned perception front-ends require. This observation is processed by a multi-stream actor--critic network (Section~\ref{subsec:policy}) trained with Proximal Policy Optimization (PPO).

To stabilize learning over long horizons, that schedule couples a five-stage curriculum (Section~\ref{subsec:curriculum}), which lengthens routes and tightens behavioral constraints stage by stage, with a stage-aware reward formulation (Section~\ref{subsec:reward}) that adapts the relative importance of mission progress, route following, safety, driving smoothness, and traffic-rule compliance in step with it. Advancing the two together is intended to keep the objectives from competing while the policy is still acquiring basic control; the factorial ablation of Section~\ref{subsec:exp_ablation} examines what the coupling is worth. In CARLA, CORAL outperforms two PPO baselines adapted from published end-to-end and curriculum-guided driving methods by a wide margin: under an identical protocol at the hardest curriculum stage, it reaches the goal in all twenty per-stage evaluation episodes while the baselines reach $5\%$ and $10\%$. A factorial ablation over the two schedules shows that neither alone matches their combination: removing either lowers both success and route completion, and disabling both drops success to $55\%$. Moreover, despite training in a single town, the learned policy transfers zero-shot: with every mission kept at the hardest setting, the final policy succeeds in $68$--$98\%$ of episodes across seven unseen towns, against $99.0\%$ in the training town, and mean lateral deviation stays below $0.35\,\mathrm{m}$. A leave-one-out test of the observation (Section~\ref{subsec:exp_lidar}) is reported alongside these results and is less favorable: on this static benchmark the LiDAR histogram can be removed without a measurable loss, which we report as a limitation of the evaluation environment rather than a property of the representation.

\section{Related Work}
\label{sec:related}

{\bf Reinforcement learning for autonomous driving:}
Reinforcement learning has attracted considerable attention in autonomous driving because it enables driving behavior to be learned through interaction with an environment rather than through manually designed control rules or expert demonstrations. Early studies applied value-based and actor--critic algorithms such as DQN~\cite{mnih2015human} and A3C~\cite{mnih2016asynchronous} to driving-related decision making from sensory observations. Later continuous-control algorithms, including DDPG~\cite{lillicrap2015continuous} and SAC~\cite{haarnoja2018soft}, made it practical to generate steering, throttle, and braking commands directly.
The development of realistic simulators such as CARLA~\cite{dosovitskiy2017carla} further accelerated RL research by providing safe and scalable environments for training and evaluation. Despite these advances, learning stable driving policies for long-horizon urban navigation remains challenging due to exploration difficulty, training instability, and the need to simultaneously satisfy multiple driving objectives.

{\bf Policy optimization for continuous control:}
PPO~\cite{schulman2017proximal} has become one of the most widely adopted RL algorithms for autonomous driving because of its stable clipped objective and effectiveness in continuous-control tasks. PPO has been applied to vehicle control in simulated driving environments, including fully end-to-end formulations that map observations directly to continuous control commands~\cite{zhao2024end}.
More recently, curriculum-guided PPO approaches such as CuRLA~\cite{uppuluri2025curla} have combined PPO with a progressively harder training curriculum. Nevertheless, many PPO-based driving systems still pair PPO with a single fixed reward, which limits their effectiveness on tasks that must balance multiple competing driving objectives.

{\bf State representation:}
State representation plays a critical role in RL-based autonomous driving because it directly affects learning efficiency, computational cost, and policy robustness. Vision-based approaches are widely used for the rich appearance and semantic content of camera imagery~\cite{bojarski2016nvidia,codevilla2018cil,chen2024end}, but they typically require deep neural networks and large training datasets while remaining sensitive to illumination changes, weather conditions, and occlusions~\cite{grigorescu2020survey}. LiDAR sensors provide geometric measurements that are largely independent of illumination.
However, processing raw point clouds using architectures such as PointNet~\cite{qi2017pointnet}, VoxelNet~\cite{zhou2018voxelnet}, and related methods introduces substantial computational overhead~\cite{guo2020pointcloud,alaba2022survey}. Alternative representations such as BEV maps reduce dimensionality but often require additional preprocessing and can discard fine geometric detail during projection~\cite{chen2020mvlidarnet}.

In addition to perception features, conditioning the policy on high-level navigational commands from a topological planner has been shown to make end-to-end driving directable toward a goal that the sensory input alone cannot disambiguate~\cite{codevilla2018cil}. Building on these observations, our agent operates on a single low-dimensional vector that pairs a compact polar LiDAR histogram with ego-frame route geometry---lateral deviation and heading alignment relative to the planned route rather than a discrete turn command---integrating perception and navigation without a learned point-cloud encoder or a rasterized BEV map.

{\bf Curriculum learning and reward design:}
Curriculum learning improves RL training by gradually increasing task difficulty, allowing agents to acquire fundamental skills before confronting more challenging objectives~\cite{bengio2009curriculum,narvekar2020curriculum}. In autonomous driving, curriculum strategies have been applied through progressive route expansion, increasing traffic density, and staged scenario difficulty~\cite{qiao2018automatically,anzalone2021reinforced}. Prior studies indicate that curriculum learning can improve exploration efficiency and policy robustness, particularly in long-horizon navigation tasks.

Reward design is equally important because autonomous driving requires balancing multiple objectives, including navigation accuracy, safety, comfort, and traffic-rule compliance. Potential-based reward shaping is a well-established technique for accelerating learning while provably leaving the optimal policy unchanged~\cite{ng1999reward}.
Early driving reward designs primarily relied on collision penalties and goal-reaching rewards, whereas more recent ones incorporate comfort-aware and risk-aware objectives~\cite{lin2020anti,wu2023risk}. Despite these advances, most existing RL-based driving systems employ fixed reward formulations throughout training. Such formulations may become suboptimal as the agent progresses from learning basic vehicle control to mastering complex navigation behaviors. Furthermore, curriculum learning and reward shaping are often investigated independently rather than as a unified training framework.

In summary, unlike CuRLA~\cite{uppuluri2025curla}, which combines a vision-based latent state, a traffic-density curriculum, and a reward that is fixed apart from a collision penalty added partway through training, our framework couples a compact polar LiDAR histogram observation and a route-distance curriculum with a reward whose component weights are scheduled across curriculum stages.
\section{Methodology}

This section presents the proposed curriculum-guided reinforcement learning framework for goal-directed autonomous driving.

\subsection{System Overview}

\begin{figure*}[!t]
\centering
\includegraphics[width=\textwidth]{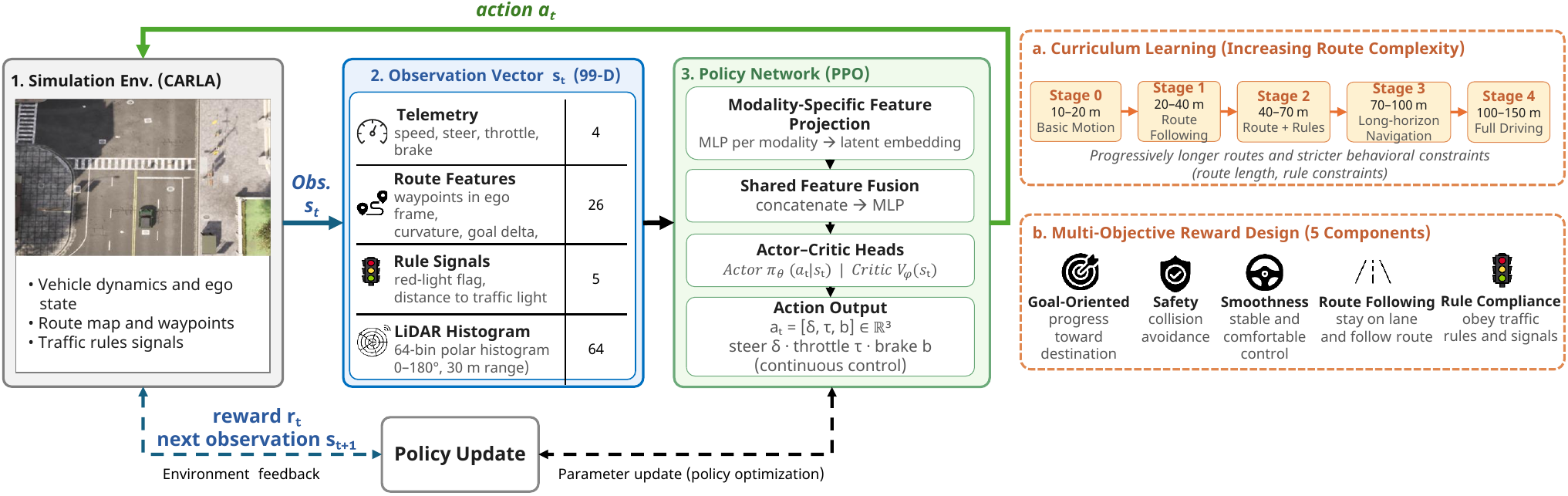}
\caption{Overview of the proposed framework: LiDAR, telemetry, route, and traffic-rule inputs form a compact observation that a multi-stream actor--critic policy maps to continuous control, trained with PPO under the coupled curriculum and stage-aware reward.}
\label{fig:system_overview}
\end{figure*}

We formulate goal-directed urban driving as a continuous-control reinforcement learning problem in which, at each timestep, the agent maps an observation of the driving scene to steering, throttle, and brake commands, and is trained in the CARLA simulator using PPO. Fig.~\ref{fig:system_overview} gives an overview of CORAL; this subsection walks through the end-to-end pipeline and points to the subsection that develops each component in detail.

At learning time, the pipeline proceeds in three steps. Raw LiDAR point clouds are first reduced to a compact polar histogram that encodes obstacle proximity in the ego-vehicle frame. This histogram is then concatenated with vehicle telemetry, route-aware navigation features, and traffic-rule awareness signals to form a single low-dimensional observation vector (Section~\ref{subsec:obs}). Finally, a multi-stream actor--critic network maps this observation to the continuous driving commands (Section~\ref{subsec:policy}).

Training couples two mechanisms within a single process. A stage-aware reward shifts the relative importance of the driving objectives (Section~\ref{subsec:reward}), while a curriculum extends route length and tightens behavioral constraints across stages (Section~\ref{subsec:curriculum}). Because the reward priorities shift in step with curriculum difficulty, the agent acquires basic vehicle control before confronting the harder navigation behaviors.

\subsection{Observation Space}
\label{subsec:obs}

\begin{figure}[t]
\centering
\includegraphics[width=\linewidth]{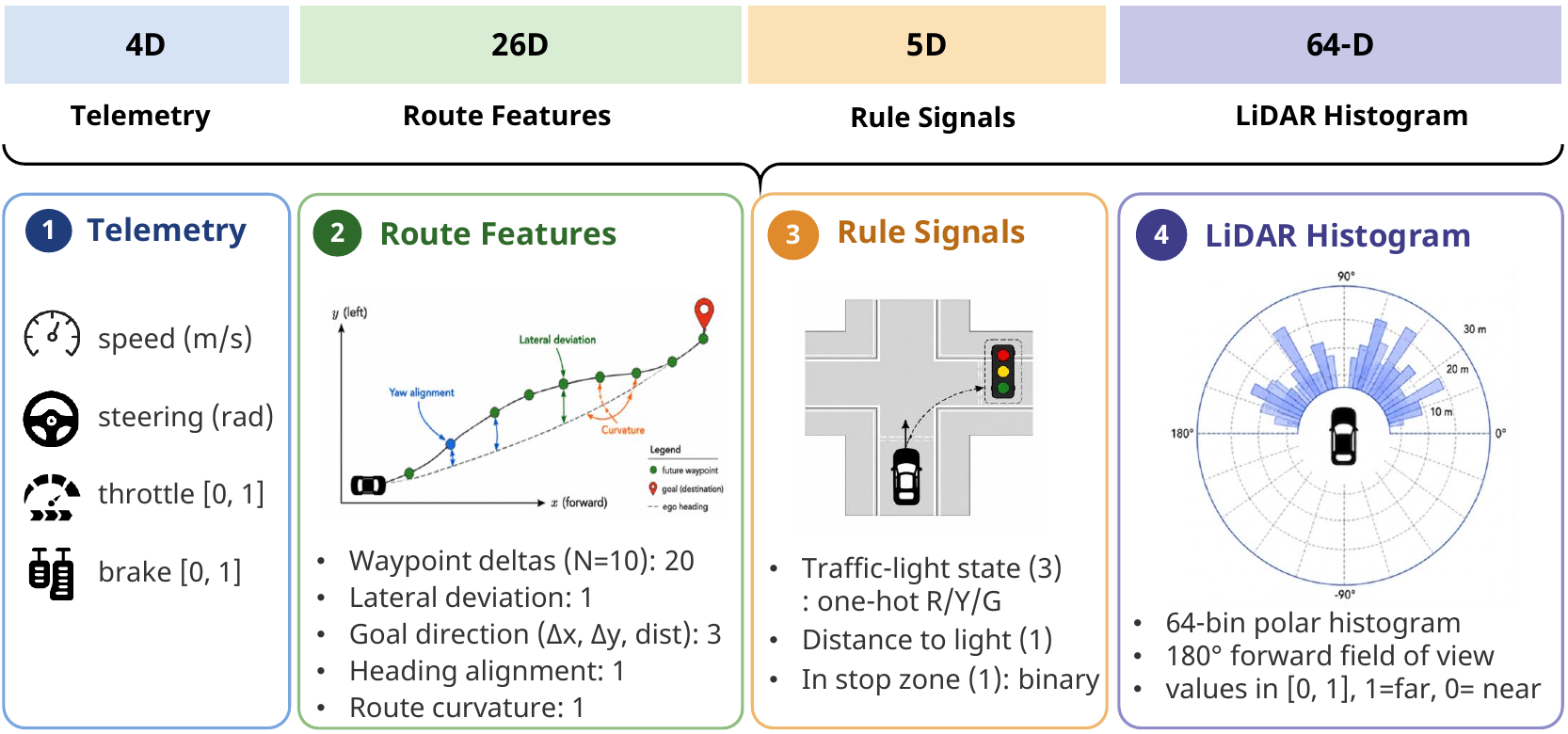}
\caption{Structure of the proposed $99$-dimensional observation vector.}
\label{fig:observation_space}
\end{figure}

The policy operates on a compact route-aware observation vector, defined at timestep $t$ as

\begin{equation}
\mathbf{o}_t=
\left[
\mathbf{o}_t^{\text{telemetry}},
\mathbf{o}_t^{\text{route}},
\mathbf{o}_t^{\text{rules}},
\mathbf{o}_t^{\text{lidar}}
\right]
\in \mathbb{R}^{99}
\label{eq:observation_vector}
\end{equation}

where the four components provide complementary information for autonomous driving. Fig.~\ref{fig:observation_space} illustrates the structure of the proposed observation representation.

Vehicle telemetry $\mathbf{o}_t^{\text{telemetry}}\in\mathbb{R}^{4}$ consists of the normalized vehicle speed, steering, throttle, and brake commands. Route-aware navigation features $\mathbf{o}_t^{\text{route}}\in\mathbb{R}^{26}$ comprise the ego-frame offsets of the next ten route waypoints ($20$ values) together with six auxiliary quantities: lateral deviation from the reference route, the goal position projected into the ego frame and the normalized distance to the goal (three goal-relative values), heading alignment with the route direction, and local route curvature. These features provide both short-term trajectory guidance and long-horizon navigation context. Traffic-rule awareness signals $\mathbf{o}_t^{\text{rules}}\in\mathbb{R}^{5}$ consist of three binary traffic-light-state indicators ($I_{\text{red}}, I_{\text{yellow}}, I_{\text{green}}$), the normalized distance to the traffic-light trigger region, and a red-light stop-zone indicator.

Local obstacle geometry is encoded as a 64-bin polar LiDAR histogram $\mathbf{o}_t^{\text{lidar}}\in\mathbb{R}^{64}$. The forward-facing $180^\circ$ field of view ($-90^\circ$ to $+90^\circ$ relative to the ego vehicle) is uniformly divided into 64 angular sectors, and each bin stores the minimum obstacle distance within its sector, normalized by a maximum sensing range of $30\,\mathrm{m}$. Smaller values therefore indicate nearby obstacles and larger values free space, yielding a compact directional encoding of obstacle proximity that avoids the cost of processing raw point clouds.

The agent controls the vehicle through a continuous action $\mathbf{a}_t=[a_t^{\text{steer}}, a_t^{\text{throttle}}, a_t^{\text{brake}}]$, where $a_t^{\text{steer}}\in[-1,1]$ is the normalized steering angle and $a_t^{\text{throttle}}, a_t^{\text{brake}}\in[0,1]$ control acceleration and braking. The continuous formulation is well suited to policy-gradient optimization with PPO and yields smoother control than discrete action sets.

\subsection{Multi-Stream Actor--Critic Policy Network}
\label{subsec:policy}

\begin{figure}[t]
\centering
\includegraphics[width=\linewidth]{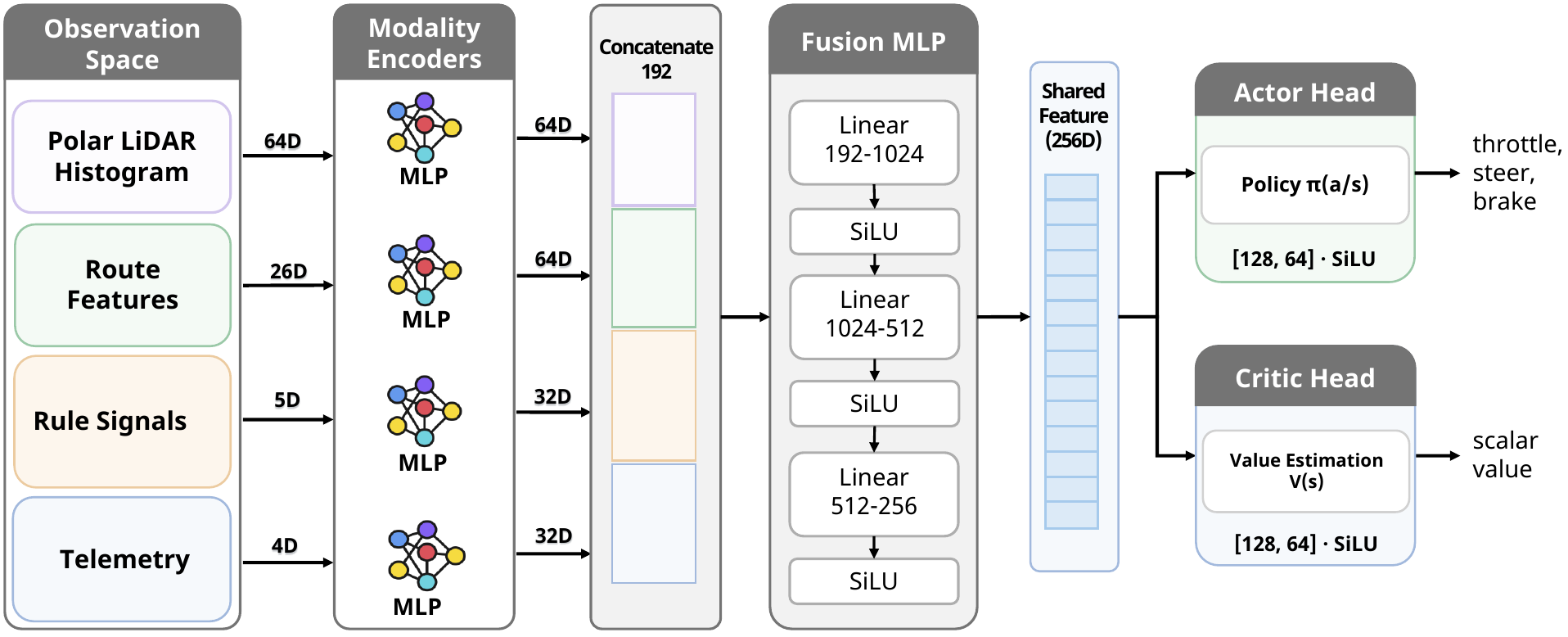}
\caption{Architecture of the proposed multi-stream actor--critic policy network.}
\label{fig:policy_network}
\end{figure}

The proposed driving policy is parameterized by a multi-stream actor--critic neural network, shown in Fig.~\ref{fig:policy_network}. The network receives the structured observation vector of Section~\ref{subsec:obs} and processes each modality through an independent encoder before feature fusion, enabling modality-specific feature extraction while reducing interference between heterogeneous inputs such as vehicle dynamics, route geometry, traffic-rule signals, and obstacle observations.

Each stream is projected into a compact latent representation by a lightweight multilayer perceptron (MLP): the telemetry ($\mathbb{R}^{4}$), route ($\mathbb{R}^{26}$), rule ($\mathbb{R}^{5}$), and LiDAR ($\mathbb{R}^{64}$) inputs are encoded into $32$-, $64$-, $32$-, and $64$-dimensional embeddings, respectively. These embeddings are concatenated into a $192$-dimensional feature and passed through a shared fusion MLP with layer widths $192\rightarrow1024\rightarrow512\rightarrow256$.

Two task-specific heads, each with hidden layers $256\rightarrow128\rightarrow64$, branch from the fused representation. The actor head parameterizes a diagonal Gaussian policy over the continuous steering, throttle, and brake commands, while the critic head estimates the state value used for advantage estimation during PPO training; the two heads share the fused features but are otherwise independent.

\subsection{Stage-Aware Reward Design}
\label{subsec:reward}

To guide policy learning in long-horizon urban navigation tasks, the proposed framework employs a multi-objective reward formulation that combines mission progress, route following, safety, smoothness, and traffic-rule objectives. Rather than relying on a single sparse reward signal, the reward is decomposed into interpretable components corresponding to distinct driving behaviors.

The reward at timestep $t$ is defined as

\begin{equation}
R_t =
w_{\text{mis}}^{(k)}R_{\text{mis}} +
w_{\text{rt}}^{(k)}R_{\text{rt}} +
w_{\text{saf}}^{(k)}R_{\text{saf}} +
w_{\text{smo}}^{(k)}R_{\text{smo}} +
w_{\text{rule}}^{(k)}R_{\text{rule}},
\label{eq:reward}
\end{equation}

where $R_{\text{mis}}$, $R_{\text{rt}}$, $R_{\text{saf}}$, $R_{\text{smo}}$, and $R_{\text{rule}}$ denote the mission, route-following, safety, smoothness, and rule-compliance rewards, respectively, and the weights $w_{\cdot}^{(k)}$ depend on the curriculum stage $k\in\{0,\dots,4\}$, allowing the learning objectives to evolve throughout training. The per-step reward is clipped to $[-10,10]$ to avoid destabilizing policy updates, and terminal events are handled separately (Eq.~\ref{eq:terminal}).

\begin{figure*}[!t]
\centering
\includegraphics[width=0.9\linewidth]{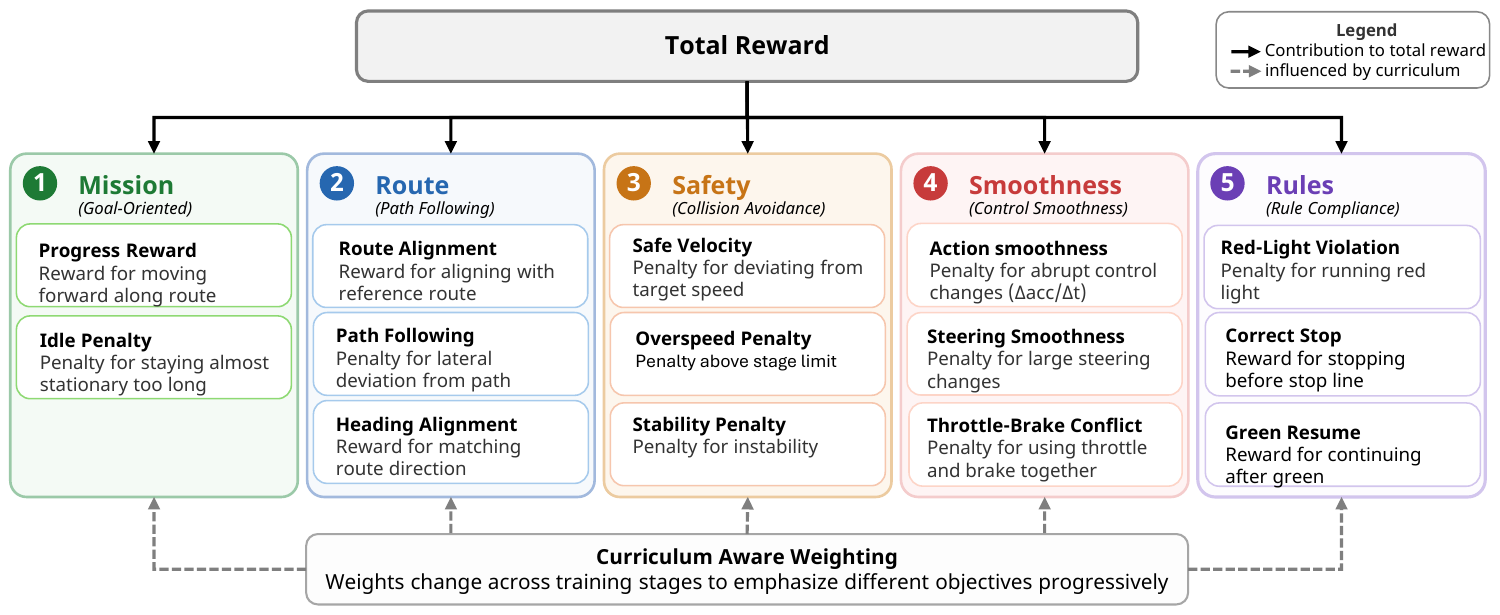}
\caption{Structure of the stage-aware reward: the five component rewards of Eq.~\ref{eq:reward} are combined with stage-dependent weights $w^{(k)}$ (Table~\ref{tab:stage_schedule}).}
\label{fig:reward_architecture}
\end{figure*}

The \emph{mission} reward encourages continuous progress toward the destination while discouraging prolonged idling. Let $\Delta d_t=d_{t-1}-d_t$ denote the reduction in the remaining route distance between consecutive timesteps. The progress increment is normalized and clipped, which bounds the contribution of unusually large transitions:

\begin{equation}
R_{\text{prog}} = \operatorname{clip}\!\left( \frac{\Delta d_t}{\bar{d}},\, -1,\, 1\right),
\end{equation}

where $\bar{d}$ is a normalization constant that scales the per-step displacement. To prevent reward accumulation while the vehicle is stationary, the progress term is suppressed when the vehicle speed falls below $0.2\,\mathrm{m/s}$. A single idle penalty, applied in the mission reward and nowhere else, discourages prolonged unnecessary stopping. An idle counter $N_t^{\text{idle}}$ is incremented whenever the speed falls below $v_{\text{idle}}=0.6\,\mathrm{m/s}$ and is reset otherwise; once it exceeds a stage-dependent patience period $N_{\text{pat}}$, the penalty is

\begin{equation}
R_{\text{idle}} =
-\operatorname{clip}\!\left(
\frac{\max(0,\,N_t^{\text{idle}}-N_{\text{pat}})}{80},\,
0,\, 1
\right),
\end{equation}

so that it grows linearly once patience is exhausted and saturates at $-1$ after a further $80$ steps ($4\,\mathrm{s}$ at $20\,\mathrm{Hz}$), giving $R_{\text{idle}}\in[-1,0]$. Here $N_{\text{pat}}=35$ steps ($1.75\,\mathrm{s}$) for Stages~0--2 and $25$ steps ($1.25\,\mathrm{s}$) for Stages~3--4. The penalty is disabled while the vehicle is legitimately stopped at a red light within the $20\,\mathrm{m}$ stop zone. Separately, in Stages~1--4, the idle penalty is disabled during an initial warm-up of $80$ steps ($4\,\mathrm{s}$). The idle-penalty saturation length and the initial warm-up are distinct implementation constants that happen to have the same value. The final mission reward is
\begin{equation}
R_{\text{mis}} = R_{\text{prog}} + R_{\text{idle}}.
\end{equation}

A terminal goal reward is assigned separately upon reaching the destination (Eq.~\ref{eq:terminal}).

The \emph{route-following} reward combines route alignment, waypoint alignment, lateral path tracking, and heading-error minimization:

\begin{align}
R_{\text{align}} &= \beta_1 A_{\text{route}} + \beta_2 \cos(\theta_{\text{wp}}), \\
R_{\text{path}} &= 1 - \frac{|e_t^{\text{lat}}|}{0.5\,W_{\text{lane}}} - \lambda_h \frac{|\theta_t^{\text{head}}|}{\pi/2}, \\
R_{\text{rt}} &= \beta_3 R_{\text{align}} + \beta_4 R_{\text{path}},
\end{align}

where $A_{\text{route}}$ is the cosine alignment between the vehicle heading and the planned route direction, $\theta_{\text{wp}}$ is the angle to the next waypoint, $e_t^{\text{lat}}$ is the lateral deviation from the reference route, $W_{\text{lane}}$ is the lane width, and $\theta_t^{\text{head}}$ is the heading error with respect to the route direction. The weights $\beta_1,\beta_2$ balance route and waypoint alignment, $\lambda_h$ trades off lateral against heading error, and $\beta_3,\beta_4$ balance the alignment and path-tracking terms.

The \emph{safety} reward regulates speed according to route geometry. The absolute local curvature is first scaled and clipped to $[0,1]$, so that the target speed reaches its minimum on the tightest curves,
\begin{equation}
\hat{\kappa}_t = \operatorname{clip}\!\left(12|\kappa_t|,\,0,\,1\right),
\end{equation}
where $\kappa_t$ is the local path curvature in $\mathrm{m}^{-1}$, estimated from consecutive route segments, so that the gain of $12\,\mathrm{m}$ saturates $\hat{\kappa}_t$ at an effective turn radius of about $12\,\mathrm{m}$: corners tighter than that are all treated as maximally curved. The target speed is then interpolated between the minimum and the stage-dependent base speed,
\begin{equation}
v_t^{\text{target}} = v_{\min} + (v_{\text{base}}-v_{\min})(1-\hat{\kappa}_t)^2,
\end{equation}
and the reward's principal term penalizes departures from it,
\begin{equation}
R_{\text{saf}} = 1 -\frac{\left|v_t-v_t^{\text{target}}\right|} {v_t^{\text{target}}}.
\end{equation}
Additional penalties discourage overspeeding and excessive combined steering and speed, while collisions are handled separately as terminal safety failures.

The \emph{smoothness} reward penalizes abrupt control changes, with $\Delta \mathbf{a}_t = \mathbf{a}_t - \mathbf{a}_{t-1}$ and steering $\delta_t$:
\begin{equation}
R_{\text{smo}} = -\alpha_{a} \left\lVert \Delta \mathbf{a}_t \right\rVert_2^2 -\alpha_{\delta} \left| \delta_t-\delta_{t-1} \right| -\alpha_{\text{tb}} \,\mathbb{I}_{\text{tb}},
\end{equation}
where $\mathbb{I}_{\text{tb}}$ equals one when throttle and brake are simultaneously applied above predefined thresholds and zero otherwise, and $\alpha_{a}$, $\alpha_{\delta}$, and $\alpha_{\text{tb}}$ weight the action-change, steering-change, and throttle--brake conflict penalties, respectively.

The \emph{rule-compliance} reward applies a large penalty $C_{\text{red}}$ for crossing a red light and positive rewards for stopping correctly and resuming on green. It is activated from Stage~2 and gradually ramped in.

Terminal events are handled separately from the weighted step reward:
\begin{equation}
R^{\text{term}} =
\begin{cases}
R_{\text{goal}}, & \text{goal reached},\\
R_{\text{crash}}, & \text{collision},\\
R_{\text{timeout}}, & \text{time limit reached},
\end{cases}
\label{eq:terminal}
\end{equation}
where the goal reward $R_{\text{goal}}=100\,s_d\,s_k$ is scaled by the initial route distance $d_0$ through $s_d=\max(0.75,\,d_0/25)$ and by the curriculum stage $k$ through $s_k=1+0.4k$, so that longer routes and later stages yield a proportionally larger completion bonus. An episode may also end when the vehicle is judged stuck, which carries no terminal reward: the idle penalty has been accumulating over the same interval already.

The scalar coefficients ($\beta_1$--$\beta_4$, $\lambda_h$, $\alpha_{a}$, $\alpha_{\delta}$, $\alpha_{\text{tb}}$, $\bar{d}$, $v_{\min}$, the per-stage $v_{\text{base}}$, and $C_{\text{red}}$), together with the terminal magnitudes ($R_{\text{crash}}$ and $R_{\text{timeout}}$), are hand-set implementation constants; their numerical values are listed in Table~\ref{tab:reward_parameters}.
Unlike conventional fixed reward formulations, the proposed framework schedules these weights across the curriculum stages, as illustrated in Fig.~\ref{fig:reward_architecture}; the per-stage values are listed with the curriculum stages in Table~\ref{tab:stage_schedule}, and the schedule itself is described in Section~\ref{subsec:curriculum}.

\begin{table*}[t]
\centering
\caption{Curriculum Stages, Learning Objectives, and Stage-Aware Reward Weights $w^{(k)}$}
\label{tab:stage_schedule}
\footnotesize
\setlength{\tabcolsep}{5pt}
\begin{tabular}{cccccccc}
\toprule
Stage $k$ & Route Distance & Learning Objective & Mission & Route Foll. & Safety & Smooth. & Rule Comp. \\
\midrule
0 & $10$--$20\,\mathrm{m}$   & Basic motion and goal seeking      & 1.00 & 0.20 & 0.60 & 0.10 & 0.00 \\
1 & $20$--$40\,\mathrm{m}$   & Route-following introduction       & 1.00 & 0.50 & 0.80 & 0.15 & 0.00 \\
2 & $40$--$70\,\mathrm{m}$   & Route adherence and rule awareness & 0.90 & 0.80 & 0.80 & 0.20 & 0.50 \\
3 & $70$--$100\,\mathrm{m}$  & Long-horizon navigation            & 0.80 & 0.90 & 0.90 & 0.25 & 0.80 \\
4 & $100$--$150\,\mathrm{m}$ & Full autonomous driving task       & 0.75 & 1.00 & 0.90 & 0.30 & 1.00 \\
\bottomrule
\end{tabular}
\end{table*}

\subsection{Curriculum-Guided Training Strategy}
\label{subsec:curriculum}

Direct training on complex routes with strict behavioral constraints often leads to unstable optimization and inefficient exploration. To address these challenges, the proposed framework employs a five-stage curriculum that progressively increases task difficulty throughout training.

The curriculum is organized according to route distance and behavioral complexity, as summarized in Table~\ref{tab:stage_schedule}. Stage~0 focuses on short-range navigation tasks ($10$--$20\,\mathrm{m}$), allowing the agent to acquire basic vehicle control and goal-directed motion. Subsequent stages gradually increase the route distance, reaching $100$--$150\,\mathrm{m}$ at Stage~4.

Beyond route length, the curriculum tightens the termination criteria themselves, listed in Table~\ref{tab:termination}: three of the five are inactive at Stage~0, and every tolerance shrinks thereafter. A sidewalk incursion that Stage~1 tolerates for $45$ steps ($2.25\,\mathrm{s}$) therefore ends the episode within $14$ steps ($0.7\,\mathrm{s}$) at Stage~4. Early stages accordingly leave room to recover from a mistake, whereas later ones hold the policy to the constraint.

\begin{table}[t]
\centering
\caption{Stage-Dependent Termination Tolerances Used During Training. Each entry is the number of consecutive simulation steps ($20\,$Hz) a violation may persist before the episode is terminated, and \textendash\ marks a criterion inactive at that stage. The evaluation protocol of Section~\ref{sec:exp_results} applies its own fixed termination criteria instead.}
\label{tab:termination}
\footnotesize
\setlength{\tabcolsep}{4pt}

\begin{tabular}{@{}lccccc@{}}
\toprule
Criterion & Stage 0 & Stage 1 & Stage 2 & Stage 3 & Stage 4 \\
\midrule
Sidewalk incursion & \textendash & 45  & 40  & 22  & 14  \\
Off-road           & 150         & 90  & 55  & 28  & 18  \\
Off-route          & \textendash & 110 & 65  & 35  & 32  \\
Opposite lane      & 120         & 80  & 45  & 22  & 14  \\
Stuck              & \textendash & \textendash & 260 & 190 & 130 \\
\bottomrule
\end{tabular}
\end{table}

Stage advancement follows a fixed interaction budget rather than success-rate gating: the agent starts in Stage~0 and enters Stages~1--4 once the global step count reaches $60{,}000$, $150{,}000$, $250{,}000$, and $350{,}000$, which allocates $60{,}000$, $90{,}000$, $100{,}000$, $100{,}000$, and $150{,}000$ of the $500{,}000$-step budget to Stages~0--4, respectively. This deterministic schedule keeps the curriculum identical across runs, at the cost of not adapting to how quickly a given run masters each stage.

This progression is tightly coupled with the stage-aware reward of Section~\ref{subsec:reward}: the route-following weight rises sharply at Stage~1 ($0.20\rightarrow0.50$) and rule-compliance weights are ramped in from Stage~2 (Table~\ref{tab:stage_schedule}). Whether advancing the two together is worth more than advancing either alone is the question the factorial ablation of Section~\ref{subsec:exp_ablation} is designed to answer. %
\section{Experiments and Results}
\label{sec:exp_results}

This section first describes the experimental setup---the simulation environment, training configuration, evaluation protocol and metrics, and baselines---and then reports the comparative and generalization results, followed by two ablation studies.

{\bf Simulation environment:}
All experiments use the CARLA simulator (v0.9.15) on a goal-directed navigation task, in which the agent drives from a randomly sampled start to a target destination while following a planned route and obeying traffic signals. The environment is \emph{static}---no other vehicles or pedestrians---so the task isolates route following and traffic-signal compliance, with collisions arising only from fixed scene geometry.
The policy is trained solely in Town05 and evaluated both there and in seven unseen towns spanning urban grids, curves, roundabouts, multi-lane roads, and complex intersections (Section~\ref{sec:exp_results}-B). In the proposed configuration the ego vehicle carries a $30\,\mathrm{m}$-range LiDAR whose point clouds form the 64-bin polar histogram of the 99-dimensional observation (Section~\ref{subsec:obs}); the two baselines instead carry the front camera their original observations require. The simulator runs at a fixed time step and the policy issues one steering--throttle--brake command per step; an episode is truncated at the step limit, which triggers the timeout penalty $R_{\text{timeout}}$. Table~\ref{tab:experimental_setup} summarizes the configuration.

\begin{table}[t]
\centering
\caption{Experimental Configuration of the Proposed Method. The baselines share the simulator, route planner, training budget, curriculum, and evaluation protocol; their sensing and observation differ as described under Baseline methods.}
\label{tab:experimental_setup}
\footnotesize

\begin{tabular}{ll}
\toprule
Parameter & Configuration \\
\midrule
Simulator & CARLA 0.9.15 \\
Training Town & Town05 \\
Evaluation Towns & Town01--Town07, Town10 \\
Observation Dimension & 99 \\
Action Space & Continuous \\
Sensor Type & LiDAR \\
LiDAR Representation & 64-bin Polar Histogram \\
RL Algorithm & PPO \\
Policy Network & Multi-Stream Actor--Critic \\
Observation Normalization & Enabled \\
Reward Normalization & Enabled \\
Curriculum Learning & Enabled \\
Simulation Time Step & $0.05$\,s ($20$\,Hz) \\
Episode Step Limit & $5000$ steps ($250$\,s) \\
Training Timesteps & 500,000 \\
Route Planner & GlobalRoutePlanner \\
Framework & Stable-Baselines3~\cite{raffin2021stable} \\
\bottomrule
\end{tabular}

\end{table}

{\bf Training configuration:}
The policy is trained with PPO under the five-stage curriculum of Section~\ref{subsec:curriculum}.
The PPO hyperparameters are listed in Table~\ref{tab:ppo_hyperparameters}, where GAE denotes generalized advantage estimation. Table~\ref{tab:reward_parameters} reports the scalar constants of the stage-aware reward of Section~\ref{subsec:reward}: the mission constants ($\bar{d}$, $v_{\text{idle}}$, $N_{\text{pat}}$, and the idle-saturation and warm-up lengths), the route weights $\beta_1$--$\beta_4$ and $\lambda_h$, the smoothness coefficients $\alpha_{a}$, $\alpha_{\delta}$, $\alpha_{\text{tb}}$, the speed bounds $v_{\min}$ and the per-stage $v_{\text{base}}$, the curvature gain, the red-light penalty $C_{\text{red}}$, and the terminal magnitudes.

\begin{table}[t]
\centering
\caption{PPO Training Hyperparameters}
\label{tab:ppo_hyperparameters}
\footnotesize
\setlength{\tabcolsep}{4pt}

\begin{tabular}{lc@{\hspace{1.8em}}|lc}
\toprule
Parameter & Value & Parameter & Value \\
\midrule
Learning Rate      & $1\times10^{-4}$ & Clip Range     & 0.2  \\
$n_{\text{steps}}$ & 2048             & Entropy Coef.  & 0.01 \\
Batch Size         & 256              & Value Coef.    & 0.5  \\
Epochs             & 8                & Max Grad Norm  & 0.5  \\
Discount $\gamma$  & 0.99             & GAE $\lambda$  & 0.95 \\
\bottomrule
\end{tabular}
\end{table}

\begin{table}[t]
\centering
\caption{Reward Function Constants, Grouped by the Reward Components of Section~\ref{subsec:reward}. Three entries depend on the curriculum stage: the five $v_{\text{base}}$ values correspond to Stages 0--4, $N_{\text{pat}}$ is $35$ steps for Stages 0--2 and $25$ for Stages 3--4, and the warm-up applies only in Stages 1--4. $R_{\text{goal}}$ additionally scales with route length and stage through $s_d$ and $s_k$ (Section~\ref{subsec:reward}); all remaining numerical constants are fixed throughout training.}
\label{tab:reward_parameters}
\footnotesize
\renewcommand{\arraystretch}{1.1}
\setlength{\tabcolsep}{3pt}

\begin{tabular}{@{}ll@{}}
\toprule
Mission &
\makecell[l]{
$\bar{d}=0.35$\,m, $v_{\text{idle}}=0.6$\,m/s,\\
$N_{\text{pat}}=35/25$ steps $(1.75/1.25$\,s$)$,\\
idle saturation $=80$ steps $(4$\,s$)$,\\
warm-up $=80$ steps $(4$\,s$)$
} \\
Route following & $(\beta_1,\dots,\beta_4)=(0.25,0.75,0.60,0.40)$, $\lambda_h=0.25$ \\
Safety &
\makecell[l]{
$v_{\min}=2.5$\,m/s, $v_{\text{base}}=\{4.0,5.0,5.5,6.0,6.5\}$\,m/s,\\
curvature gain $=12$\,m
} \\
Smoothness & $(\alpha_{a},\alpha_{\delta},\alpha_{\text{tb}})=(0.08,0.05,0.30)$ \\
Rule compliance & $C_{\text{red}}=-8.0$ \\
Terminal & $R_{\text{goal}}=100\,s_d\,s_k$, $R_{\text{crash}}=-60$, $R_{\text{timeout}}=-10$ \\
\bottomrule
\end{tabular}
\end{table}

{\bf Evaluation protocol and metrics:}
The two main evaluations---the per-stage comparison and the cross-town transfer---differ both in what is held fixed and in when they are measured. The \emph{per-stage} comparison against the baselines (Section~\ref{sec:exp_results}-A) stays in the training town, Town05, and sweeps the five curriculum stages, so route length and behavioral difficulty vary while the map does not; each of its cells aggregates $20$ episodes recorded \emph{during training} at that stage, and it therefore describes how each system behaves while it is acquiring the task. The stage boundaries of Section~\ref{subsec:curriculum} delimit the same training-step windows for all three systems, so a given stage indexes the same interval of training in every row. The \emph{cross-town} evaluation (Section~\ref{sec:exp_results}-B) reverses what the per-stage comparison holds fixed: it fixes the mission at the Stage-4 setting---$100$--$150\,\mathrm{m}$ routes---and varies the map, running $100$ randomly sampled start--goal episodes in each of the eight towns.
Every post-training evaluation---the cross-town runs and the two ablations of Sections~\ref{subsec:exp_ablation} and~\ref{subsec:exp_lidar}---evaluates the final checkpoint at the end of the $500{,}000$-step budget, acting deterministically and without exploration noise. The post-training evaluations differ in how their episodes are drawn. The cross-town evaluation samples its $100$ episodes independently in each town, since the towns share no common route set. Each ablation instead draws one list of $100$ Stage-4 Town05 start--goal pairs and replays it identically across the configurations it compares, so that differences within a table are differences between policies rather than between routes; the two ablations use separate lists, and no comparison is made across them.

Evaluation episodes terminate under one of four conditions: (i) reaching the destination, (ii) collision, (iii) stuck-vehicle termination, or (iv) timeout. A stuck termination is triggered when the vehicle remains below $0.15\,\mathrm{m/s}$ for $600$ consecutive simulation steps ($30\,\mathrm{s}$) while not legitimately waiting at a red traffic light. Episodes that reach the maximum duration of $5000$ simulation steps ($250\,\mathrm{s}$) without satisfying another termination condition are classified as timeouts.
Seven complementary metrics measure navigation success, safety, route-following accuracy, and traffic-rule compliance:

\begin{itemize}
\item $P_{\text{succ}}$ (\%): \textbf{success rate} --- episodes in which the vehicle came within $3.0\,\mathrm{m}$ of the final route waypoint.
\item $P_{\text{coll}}$ (\%): \textbf{collision rate} --- episodes terminated by a collision.
\item $P_{\text{off}}$ (\%): \textbf{off-route rate} --- percentage of evaluation timesteps at which the vehicle deviates excessively from the reference route, judged by the route planner's adaptive distance criterion over the local lane width, route curvature, curriculum stage, and junctions. Moderate deviations count only after a stage-dependent patience threshold, whereas deviations beyond $1.5\times$ the lane width count immediately. Because the criterion depends on the curriculum stage, $P_{\text{off}}$ compares methods at a given stage rather than one stage against another; the towns of Table~\ref{tab:generalization_results} are all evaluated at the Stage-4 setting and so are mutually comparable.
\item $\bar{e}_{\text{lat}}$ (m): \textbf{lateral deviation} --- mean perpendicular distance to the reference route, averaging the per-step $e_t^{\text{lat}}$ of Section~\ref{subsec:reward}.
\item $P_{\text{lane}}$ (\%): \textbf{lane-invasion rate} --- number of detected lane-boundary invasion events normalized by the total number of evaluation timesteps.
\item Red Lights (E / V): \textbf{red-light encounters and violations} --- $E$ counts encounters and $V$ violations. A violation is recorded when the vehicle crosses the stop line and enters an intersection while the corresponding traffic signal remains red.
\item $P_{\text{comp}}$ (\%): \textbf{route completion} --- $1-d_{\text{final}}/d_0$, the fraction of the initial route distance $d_0$ that has been covered when the episode ends at remaining distance $d_{\text{final}}$. Unlike $P_{\text{succ}}$, which scores every unsuccessful episode alike, it gives partial credit on episodes that never reach the goal, and it is therefore reported only in the ablation of Section~\ref{subsec:exp_ablation}, where several configurations fail outright.
\end{itemize}

Higher is better for $P_{\text{succ}}$ and $P_{\text{comp}}$ and lower for the remaining metrics; for Red Lights, $E$ provides exposure context, so a lower $V$ is preferable at a comparable $E$.

\smallskip
{\bf Baseline methods:}
The framework is compared against two PPO-based autonomous driving baselines.

\begin{itemize}
    \item \textbf{E2E PPO} follows the end-to-end reinforcement learning framework of~\cite{zhao2024end}, which maps observations directly to continuous vehicle control actions using PPO under a reward centered on lane keeping, heading alignment, speed regulation, and collision avoidance. Its original observation is retained---front-camera images encoded by a pre-trained variational autoencoder (VAE), together with auxiliary vehicle states---and augmented with the complete $26$-dimensional route block of Section~\ref{subsec:obs} so that it addresses the same goal-directed task. It receives neither the polar LiDAR histogram nor the traffic-rule signals; it is trained under the same route-distance curriculum as the other systems, with its reward left unchanged across stages.

    \item \textbf{CuRLA-Inspired PPO} is derived from CuRLA~\cite{uppuluri2025curla}, a curriculum-guided PPO method. Because the environment used here contains no dynamic traffic participants, CuRLA's traffic-density curriculum is not directly applicable; this baseline therefore trains under the same route-distance curriculum as the other two systems, and what remains specifically CuRLA's is its observation and reward. That observation is retained---a VAE-based visual representation with external state variables---and augmented with the same $26$-dimensional route block; it receives neither the polar LiDAR histogram nor the traffic-rule signals, and its reward is CuRLA's own formulation with none of the stage-aware reweighting applied. It should thus be read as an adaptation rather than a faithful reimplementation of CuRLA.
\end{itemize}

For fairness, all three systems shared the same CARLA environment, PPO implementation, route planner, training budget, evaluation protocol, progressive route-distance curriculum, and $26$-dimensional route observation. What the comparison varies is therefore the rest of the observation---the polar LiDAR histogram and the traffic-rule signals, which only the proposed method receives, against the VAE-encoded camera stream, which only the baselines do---and the reward: stage-aware weights against each baseline's own fixed formulation. Section~\ref{subsec:exp_ablation} isolates the training-schedule factors and Section~\ref{subsec:exp_lidar} the LiDAR stream.

\subsection{Comparative Performance Evaluation}

\begin{table*}[!t]
\centering
\caption{Per-Stage Comparison Against the PPO Baselines in the Training Town (Town05). Each of the three systems is a single training run. Each cell aggregates $20$ episodes recorded \emph{during training} at that stage ($10$ at its midpoint, $10$ after its completion); metrics are defined in Section~\ref{sec:exp_results}, stage difficulty in Table~\ref{tab:stage_schedule}. Boldface marks the proposed method, not the best value in each column.}
\label{tab:comparative_results}
\footnotesize
\setlength{\tabcolsep}{3.5pt}

\begin{tabular}{llcccccc}
\toprule
Method & Stage (route distance) & $P_{\text{succ}}$ (\%) & $P_{\text{coll}}$ (\%) & $P_{\text{off}}$ (\%) & $\bar{e}_{\text{lat}}$ (m) & $P_{\text{lane}}$ (\%) & Red Lights (E / V) \\
\midrule
\multirow{5}{*}{E2E PPO~\cite{zhao2024end}}
 & 0 ($10$--$20$\,m)   & 10.0 & 20.0 & 38.47 & 0.52 & 2.76 & --- \\
 & 1 ($20$--$40$\,m)   & 20.0 & 40.0 & 30.56 & 0.67 & 4.92 & --- \\
 & 2 ($40$--$70$\,m)   & 25.0 & 50.0 & 13.93 & 0.69 & 5.00 & --- \\
 & 3 ($70$--$100$\,m)  & 25.0 & 60.0 & 10.56 & 0.76 & 7.57 & --- \\
 & 4 ($100$--$150$\,m) & 5.0  & 80.0 & 12.13 & 0.80 & 3.87 & --- \\
\midrule
\multirow{5}{*}{CuRLA-Inspired PPO~\cite{uppuluri2025curla}}
 & 0 ($10$--$20$\,m)   & 100.0 & 0.0 & 0.00 & 0.13 & 1.03 & --- \\
 & 1 ($20$--$40$\,m)   & 80.0  & 20.0 & 8.29 & 0.69 & 5.29 & --- \\
 & 2 ($40$--$70$\,m)   & 30.0  & 70.0 & 20.22 & 0.72 & 8.33 & --- \\
 & 3 ($70$--$100$\,m)  & 20.0  & 80.0 & 19.37 & 1.00 & 9.63 & --- \\
 & 4 ($100$--$150$\,m) & 10.0  & 75.0 & 19.49 & 0.95 & 9.40 & --- \\
\midrule
\multirow{5}{*}{\textbf{CORAL} (ours)}
 & 0 ($10$--$20$\,m)   & \textbf{100.0} & \textbf{0.0} & \textbf{0.00} & \textbf{0.10} & \textbf{0.33} & 0/0 \\
 & 1 ($20$--$40$\,m)   & \textbf{100.0} & \textbf{0.0} & \textbf{0.00} & \textbf{0.23} & \textbf{0.61} & 3/3 \\
 & 2 ($40$--$70$\,m)   & \textbf{100.0} & \textbf{0.0} & \textbf{0.18} & \textbf{0.27} & \textbf{0.68} & 21/15 \\
 & 3 ($70$--$100$\,m)  & \textbf{95.0}  & \textbf{5.0} & \textbf{0.08} & \textbf{0.27} & \textbf{0.67} & 24/14 \\
 & 4 ($100$--$150$\,m) & \textbf{100.0} & \textbf{0.0} & \textbf{0.59} & \textbf{0.19} & \textbf{0.36} & 30/14 \\
\bottomrule
\end{tabular}
\end{table*}

Table~\ref{tab:comparative_results} compares the proposed framework against the two PPO baselines across the five curriculum stages, all three trained and evaluated in the training town, Town05.

Because the stages form a difficulty ladder, a method that merely drives competently is not expected to hold its success rate across them. Whether a system holds up under that ladder is precisely what separates the three: the proposed policy stays at $100\%$ in every stage but Stage~3 ($95\%$), whereas both baselines lose most of their success as the horizon grows. The flatness of the proposed policy's success rate across stages, rather than its value at any single stage, is the result of interest.

Success and collision rates need not be complementary, because collision is not the only unsuccessful outcome: episodes also end by timeout or stuck termination (Section~\ref{sec:exp_results}), so a method can combine a low collision rate with a low success rate.

The per-stage breakdown also localizes \emph{where} the baselines fail, which a single aggregate number would hide. All three systems were given the same route guidance to follow, as a sequence of waypoints from the same planner, so the gap is not one of being told where to go. On the route information the difference is not in what reaches the policy but in what reaches its objective: all three receive the same route block, alignment, lateral error and curvature included, yet only the proposed method's reward is defined against the planned route---the baselines score heading and lateral offset relative to the lane they occupy, not to the route they were given---so route following is something only that method is optimized for rather than merely pointed at. CuRLA-Inspired PPO completes every Stage~0 episode, demonstrating that its policy is capable of basic lane-level driving. However, its success rate drops to $30.0\%$ at Stage~2 and $10.0\%$ at Stage~4, indicating increasing difficulty in maintaining reliable route-following over longer navigation tasks. E2E PPO exhibits a different failure pattern. Its success rate is non-monotonic across the curriculum, increasing from $10.0\%$ at Stage~0 to $25.0\%$ at Stages~2 and~3 before dropping to $5.0\%$ at Stage~4. Examination of the evaluation logs shows that the poor Stage~0 performance is not primarily caused by collisions. Only four of the twenty evaluation episodes ended in collision, whereas the remaining unsuccessful episodes consisted of eight stuck terminations and six timeouts. These results indicate that the policy frequently failed to make sufficient progress toward the destination despite avoiding collisions. As the route length increased, the policy occasionally completed the assigned route, but collision failures became increasingly dominant, resulting in the sharp performance degradation observed at Stage~4.

On the remaining metrics the proposed policy is at or below both baselines at every stage, but the two baselines do not keep a fixed order between themselves: E2E PPO is the weaker of the two at the shortest routes and CuRLA-Inspired PPO from the middle stages onward on off-route rate, lane invasion and lateral deviation, while on collision rate the two exchange places twice. The proposed policy collides in one of the twenty Stage-3 episodes and in none of the others, and stays below $0.6\%$ off-route at every stage, whereas from Stage~1 onward both baselines spend between $8\%$ and $31\%$ of their timesteps away from the planned route.
Route-following accuracy separates the three systems by a factor of roughly three: averaged over the five stages, lateral deviation is $0.21\,\mathrm{m}$ for the proposed policy against $0.69\,\mathrm{m}$ for E2E PPO and $0.70\,\mathrm{m}$ for CuRLA-Inspired PPO, and it never exceeds $0.27\,\mathrm{m}$ at any stage while the baselines reach $0.80$ and $1.00\,\mathrm{m}$. This is the narrowest of the margins in the table---collision and off-route rates separate the systems by more than an order of magnitude---but it is the one that speaks most directly to the mechanism: it is the quantity the stage-aware reward penalizes directly, as deviation from the planned route, whereas both baseline rewards penalize lateral deviation only from the lane center. It also explains why the baselines can stay on a road without staying on the \emph{route}.

Red-light behavior needs both counts to interpret, because a low violation count is equally consistent with a policy that stops reliably and one that rarely meets a red light. Traffic-light compliance enters the reward only from Stage~2 and is ramped in from there, and the counts follow that schedule: at Stage~1, before the term is active, all three encountered red lights are run, whereas from Stage~2 onward the violation share falls with continued exposure---$15$ of $21$ encounters at Stage~2, $14$ of $24$ at Stage~3, and $14$ of $30$ at Stage~4---so the encounters rise by nearly half while the violations do not rise at all. Because these values are recorded during training, they trace the acquisition of the behavior rather than its final quality.

\begin{figure}[t]
\centering
\newcommand{\tlcell}[2]{%
  \begin{minipage}[t]{0.485\linewidth}\centering
    \includegraphics[width=\linewidth]{Figures/traffic_light_fig/#1}\\
    {\footnotesize #2}
  \end{minipage}}
\tlcell{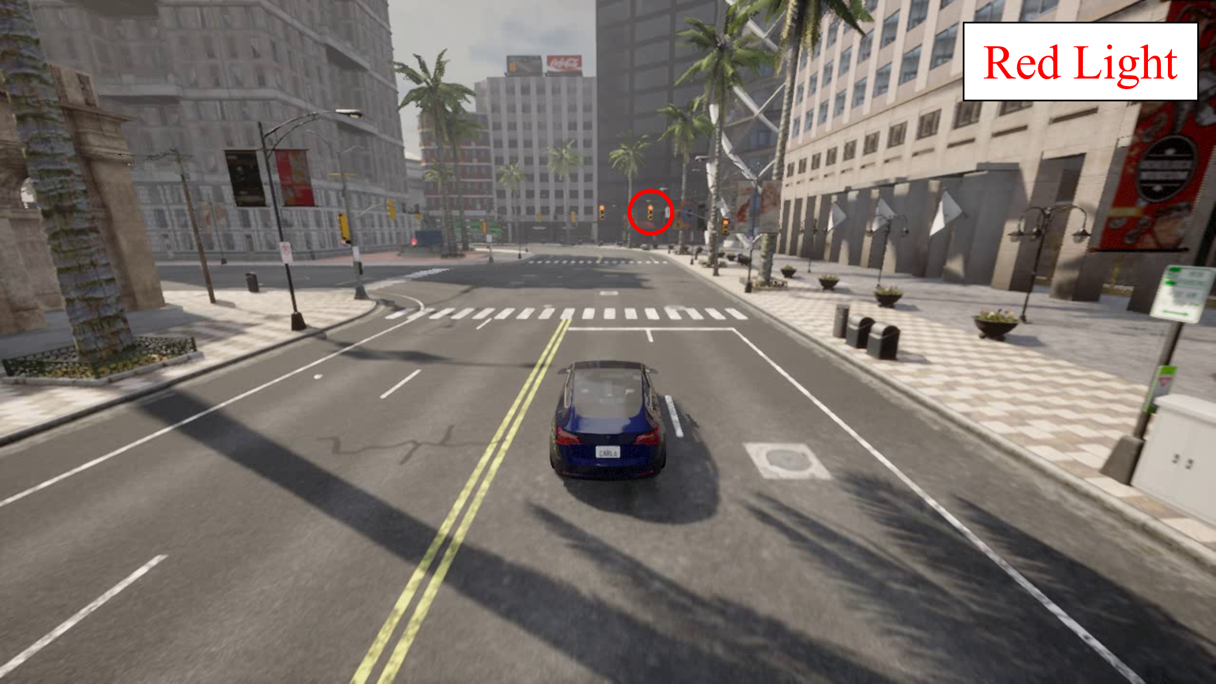}{(a) Approaching on red}\hfill
\tlcell{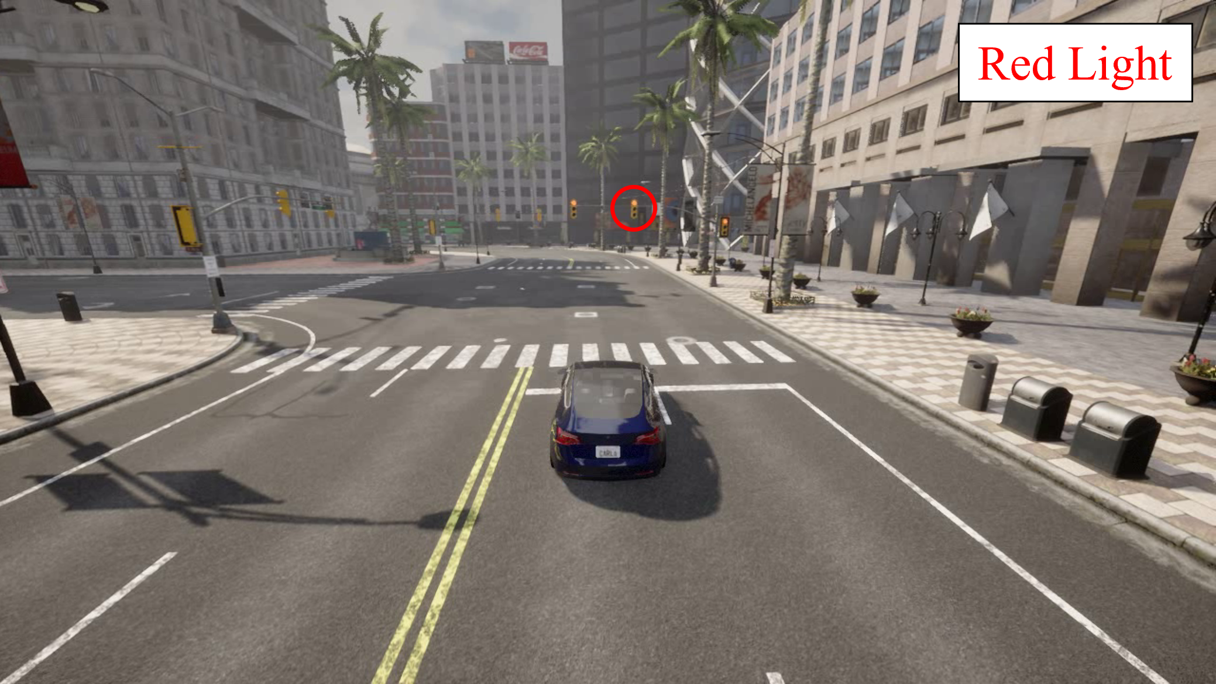}{(b) Held at the stop line}

\vspace{5pt}

\tlcell{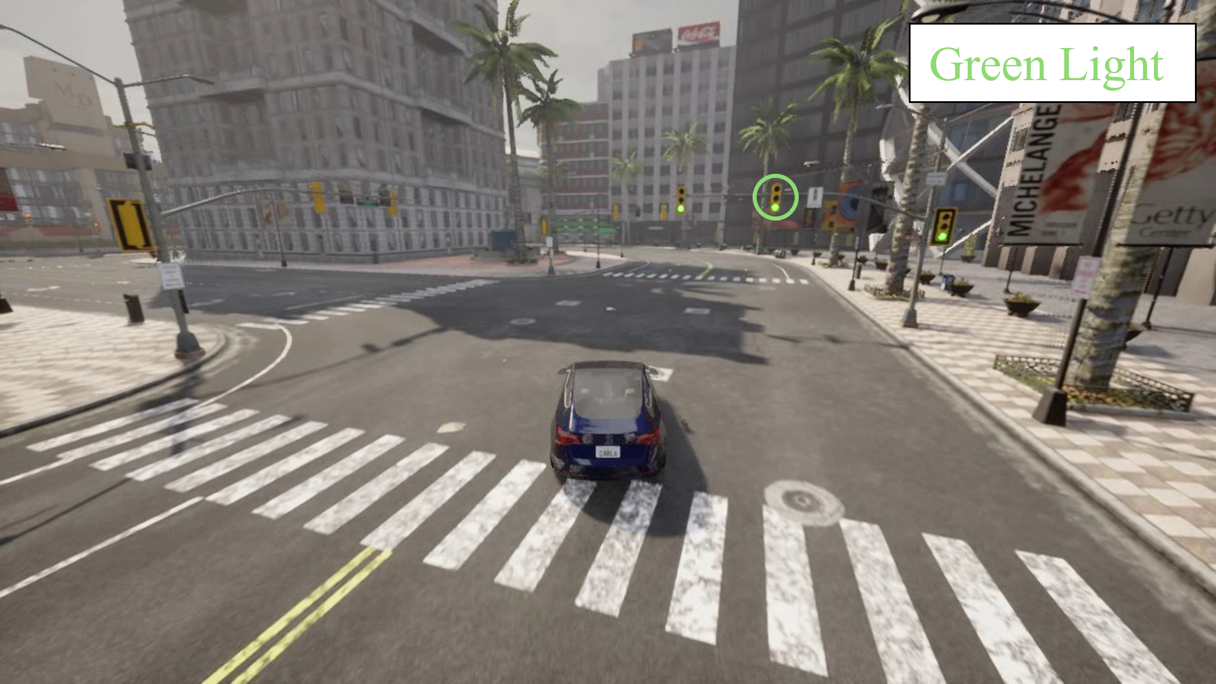}{(c) Resuming on green}\hfill
\tlcell{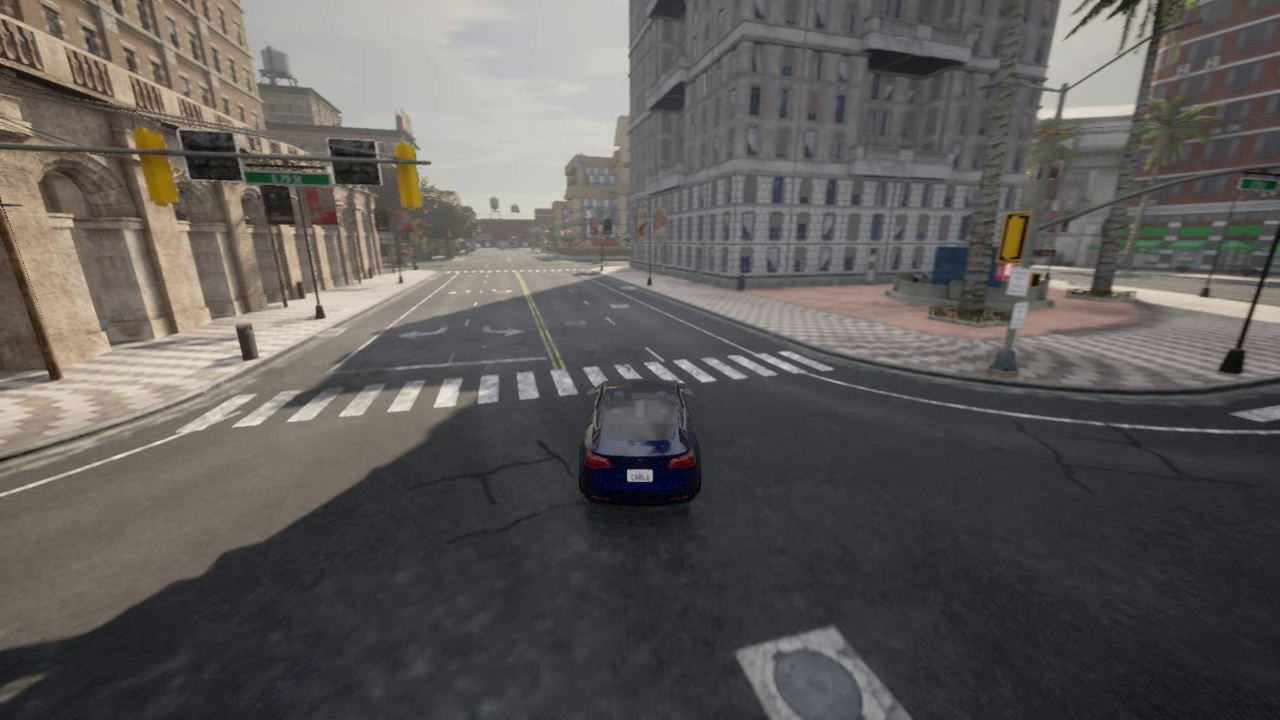}{(d) Clear of the intersection}

\caption{Traffic-light-compliant behavior of the proposed policy, in temporal order. The signal governing the ego vehicle is circled in the first three frames; in (d) the intersection has been cleared.}
\label{fig:traffic_light}
\end{figure}

The deterministic evaluation gives the latter. Across the eight towns of Table~\ref{tab:generalization_results} the fully trained policy encounters $986$ red lights and violates $347$ of them, $35\%$ overall, including $71$ of $176$ in the training town itself. Fig.~\ref{fig:traffic_light} shows the compliant behavior the policy has acquired, but the aggregate counts show that it does not yet apply it reliably. Red-light compliance is accordingly the behavior that most clearly separates what the stage-aware reward has improved from what it has solved. Neither baseline optimizes traffic-light compliance in its reward, so the column is left empty for them and no comparison is claimed on it.

\subsection{Generalization Performance}

\begin{figure*}[!t]
\centering
\newcommand{\towncell}[2]{%
  \begin{minipage}[t]{0.235\textwidth}\centering
    \includegraphics[width=0.8\linewidth]{Figures/Towns/#1.png}\\
    {\footnotesize #2}
  \end{minipage}}
\towncell{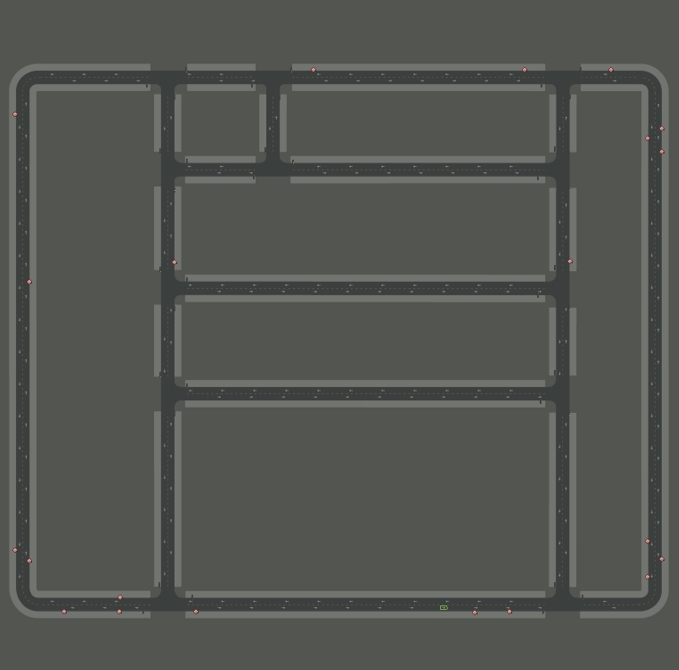}{Town01}\hfill
\towncell{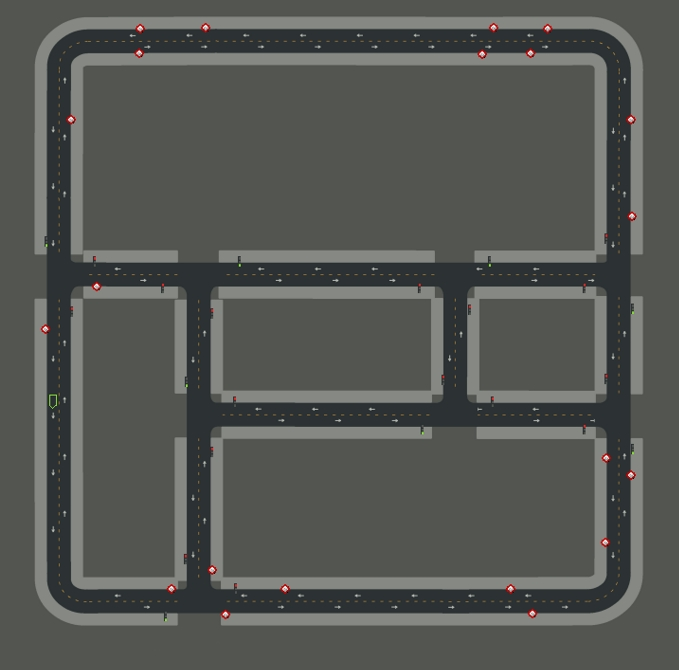}{Town02}\hfill
\towncell{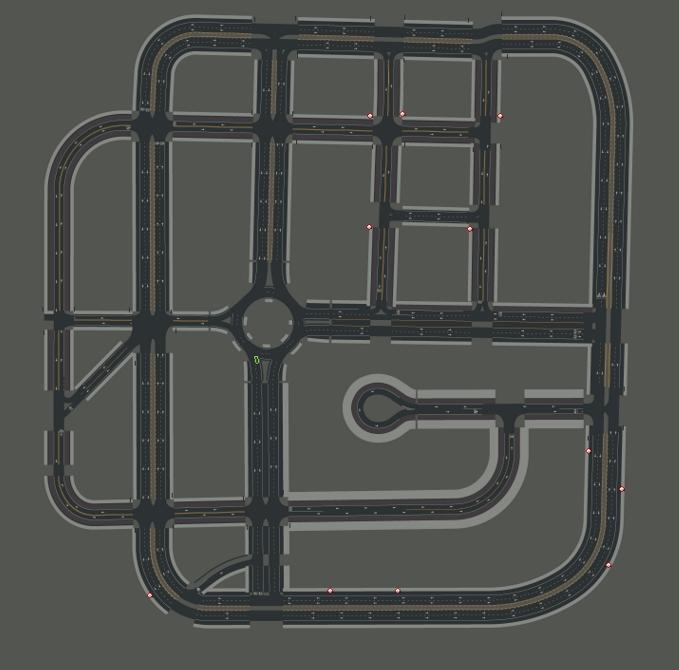}{Town03}\hfill
\towncell{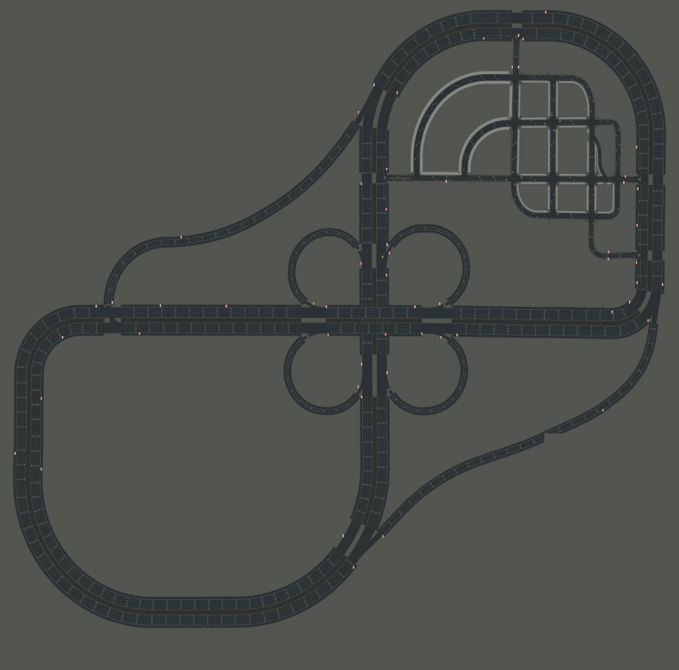}{Town04}

\vspace{6pt}

\towncell{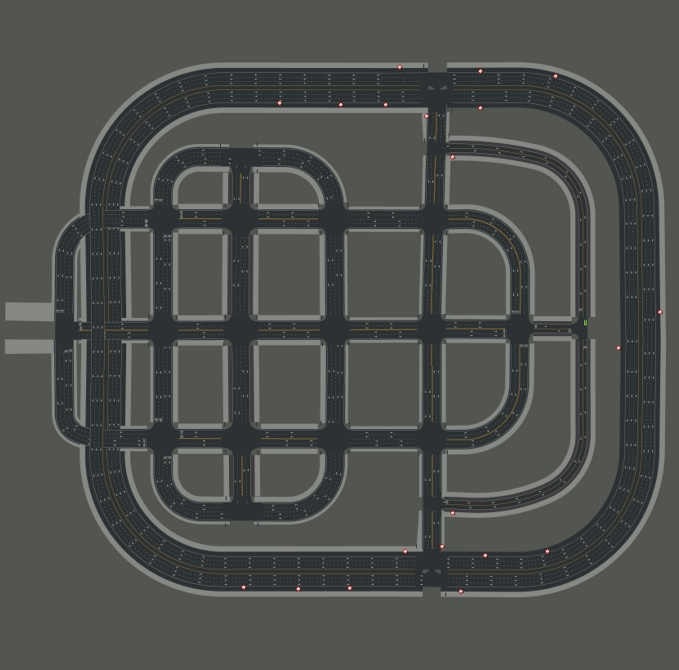}{\textbf{Town05}}\hfill
\towncell{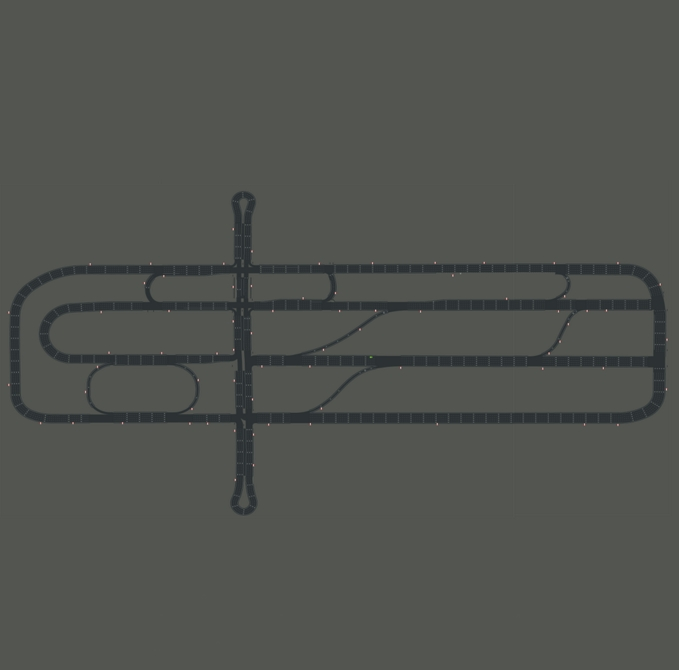}{Town06}\hfill
\towncell{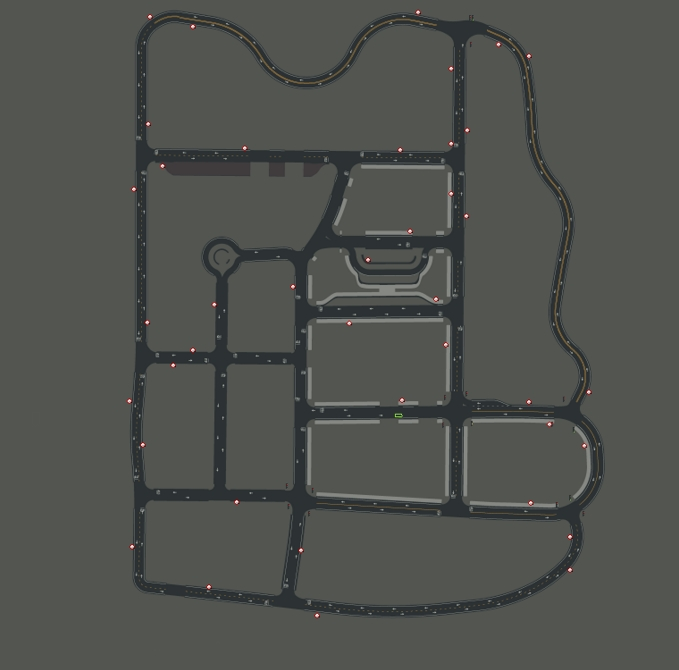}{Town07}\hfill
\towncell{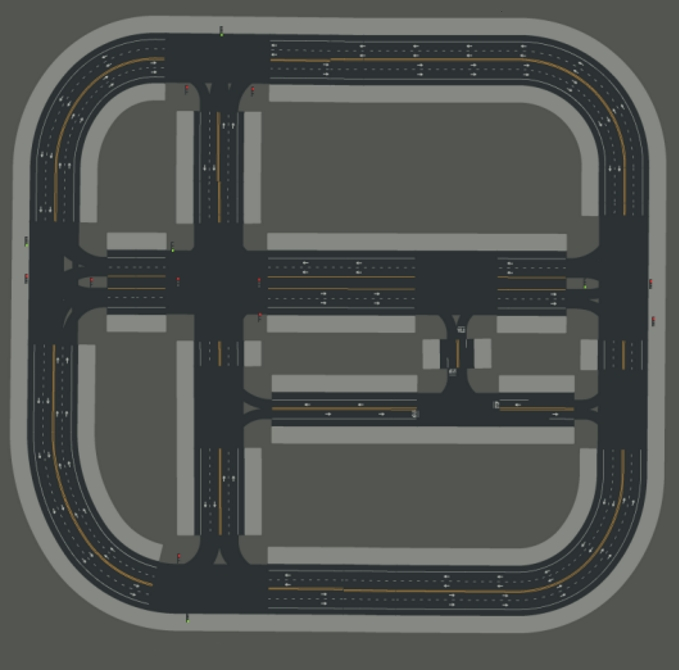}{Town10}

\caption{Top-down layouts of the eight CARLA~\cite{dosovitskiy2017carla} towns used for evaluation, each scaled individually to a common box. The policy is trained only in Town05 (bold); the other seven are unseen at training time.}
\label{fig:towns}
\end{figure*}

To test transfer, the trained policy was evaluated in eight CARLA towns---the training environment (Town05) and seven previously unseen ones (Town01--Town04, Town06, Town07, Town10). Fig.~\ref{fig:towns} shows their layouts; Table~\ref{tab:generalization_results} summarizes the resulting performance.

\begin{table}[t]
\centering
\caption{Transfer Across the Eight Evaluation Towns, With Every Mission Fixed at the Stage-4 Setting ($100$--$150$\,m Routes) and $100$ Episodes per Town. Town05 is the training town; the remaining seven are unseen. Metrics are defined in Section~\ref{sec:exp_results}; rates are percentages, $\bar{e}_{\text{lat}}$ is in meters, and the last column gives red-light encounters and violations as counts.}
\label{tab:generalization_results}
\footnotesize
\setlength{\tabcolsep}{2.5pt}

\begin{tabular}{lcccccc}
\toprule
Town & $P_{\text{succ}}$ & $P_{\text{coll}}$ & $P_{\text{off}}$ & $\bar{e}_{\text{lat}}$ & $P_{\text{lane}}$ & Red (E\,/\,V) \\
\midrule
Town01 & 93.0 & 7.0 & 0.17 & 0.23 & 0.41 & 121/55 \\
Town02 & 83.0 & 16.0 & 0.31 & 0.27 & 0.46 & 185/28 \\
Town03 & 94.0 & 5.0 & 1.65 & 0.28 & 0.61 & 111/58 \\
Town04 & 68.0 & 25.0 & 2.47 & 0.26 & 0.70 & 179/59 \\
Town05 & 99.0 & 1.0 & 0.10 & 0.19 & 0.37 & 176/71 \\
Town06 & 98.0 & 2.0 & 0.04 & 0.17 & 1.37 & 24/7 \\
Town07 & 78.0 & 22.0 & 2.11 & 0.34 & 1.44 & 77/31 \\
Town10 & 97.0 & 3.0 & 0.83 & 0.32 & 0.64 & 113/38 \\
\bottomrule
\end{tabular}
\end{table}

The policy transfers to layouts it never trained on: success ranges from $68.0\%$ to $99.0\%$ across the eight towns, reaching $99.0\%$ in the training environment Town05 and staying at or above $93.0\%$ in Town01, Town03, Town06 and Town10. Route-following accuracy is preserved throughout: mean lateral deviation remains below $0.35\,\mathrm{m}$ in every town and lane-invasion rates below $1.5\%$, while the red-light encounter and violation counts show that traffic-signal compliance varies across maps and exposure levels.

Collisions, however, are the dominant failure mode rather than a negligible one. The collision rate ranges from $1.0\%$ in Town05 to $25.0\%$ in Town04, reaching $16.0\%$ in Town02 and $22.0\%$ in Town07. In five of the eight towns it is exactly the complement of the success rate, indicating that every unsuccessful episode there terminates through collision with a static obstacle, since the environment contains no other traffic. Town02, Town03 and Town04 are the exceptions, where the remaining $1\%$, $1\%$ and $7\%$ of episodes terminate due to timeout.

Difficulty is not uniform across the evaluated towns. Town04 is the most challenging environment, with the lowest success rate ($68.0\%$), the highest collision rate ($25.0\%$) and the highest off-route rate ($2.47\%$); Town07 is next, at $78.0\%$ success, $22.0\%$ collisions and $2.11\%$ off-route. Off-route behavior concentrates in three towns---Town04, Town07 and Town03, at $2.47\%$, $2.11\%$ and $1.65\%$---while the remaining five stay at or below $0.83\%$. The two rankings do not coincide, however: Town03 is third on off-route yet among the strongest on success ($94.0\%$), and Town02 is third worst on success ($83.0\%$) with an off-route rate of only $0.31\%$. Failing often and wandering off route are therefore separate phenomena in this evaluation. The layouts differ visibly in geometry and intersection type (Fig.~\ref{fig:towns}), but this paper defines no measure of layout dissimilarity, so the spread is not attributed to a structural cause.

That a policy trained in one town holds at or above $68\%$ success in all seven unseen towns indicates that it has learned transferable behavior rather than memorizing a single layout. Which part of the design produces that transfer is not resolved here: the ablations of Sections~\ref{subsec:exp_ablation} and~\ref{subsec:exp_lidar} vary one factor at a time but do so entirely within Town05, and the baselines were never run cross-town.

\subsection{Ablation: Reward and Distance Schedules}
\label{subsec:exp_ablation}

This subsection isolates the two halves of the training schedule, which Section~\ref{sec:exp_results}-A cannot separate: the distance schedule is shared with both baselines there, and the reward schedule varies only as one part of an integrated system. Section~\ref{subsec:curriculum} couples two mechanisms: a \emph{reward schedule}, which advances the per-stage reward weights of Table~\ref{tab:stage_schedule}, and a \emph{distance schedule}, which expands the sampled route distance. Because the claim is that the two should advance \emph{together}, removing one at a time is not enough---that measures only what each is worth in the presence of the other, never what either does alone. We therefore run the complete $2\times2$ factorial, whose fourth cell---neither schedule---supplies the missing comparison.

Table~\ref{tab:curriculum_ablation} lists the four configurations: the full schedule, which is the proposed method; the two single-factor variants; and a no-curriculum control. Turning a schedule off means holding everything it controls at its Stage-4 value for the whole run---the stage-dependent terms of the reward function in one case, the sampled route distance and the behavioral constraints that tighten with it in the other---so all four runs are asked to solve the same terminal task and differ only in how they are led to it. The terminal goal reward continues to scale with the sampled initial route length through the distance factor $s_d$ (Section~\ref{subsec:reward}), so configurations trained on longer routes receive proportionally larger terminal rewards even when the reward schedule is held fixed. All share the network of Section~\ref{subsec:policy}, the observation of Section~\ref{subsec:obs}, the hyperparameters of Table~\ref{tab:ppo_hyperparameters}, the constants of Table~\ref{tab:reward_parameters}, and a $500{,}000$-step budget, which fixes the number of PPO updates identically across cells, though not the number of episodes: long routes consume more steps than short ones, so the cells whose distance schedule is active complete more of them. The two contrasts that isolate the reward schedule hold the route-length distribution fixed and are unaffected by this.

\begin{table}[t]
\centering
\caption{Factorial Ablation of the Training Schedule. Each configuration is a single training run, evaluated with its final checkpoint on one list of $100$ Stage-4 Town05 start--goal pairs replayed identically across the four. \checkmark: the schedule advances through the curriculum stages; \textendash: held at its Stage-4 setting throughout. $P_{\text{comp}}$: route completion (Section~\ref{sec:exp_results}).}
\label{tab:curriculum_ablation}
\footnotesize
\setlength{\tabcolsep}{4pt}
\renewcommand{\arraystretch}{1.1}

\begin{tabular}{@{}lcccc@{}}
\toprule
 & \multicolumn{2}{c}{Curriculum} & & \\
\cmidrule(lr){2-3}
Configuration & Reward & Distance & $P_{\text{succ}}$ (\%) & $P_{\text{comp}}$ (\%) \\
\midrule
Full (CORAL) & \checkmark & \checkmark & 99.0 & 96.40 \\
Distance only   & \textendash & \checkmark & 72.0 & 89.82 \\
Reward only     & \checkmark & \textendash & 90.0 & 90.19 \\
No curriculum   & \textendash & \textendash & 55.0 & 87.10 \\
\bottomrule
\end{tabular}
\end{table}

Two scores are reported because success alone cannot carry the comparison: success measures how often the goal is reached, while route completion, defined for every episode (Section~\ref{sec:exp_results}), gives partial credit for how much of the route is covered when it is not.

The full schedule is the best of the four on both scores, at $99.0\%$ success and $96.40\%$ completion, and removing either schedule costs on both. The two removals are not equally expensive on success---dropping the reward schedule costs $27.0$ points against the distance schedule's $9.0$---but on completion they cost almost the same, $6.58$ and $6.21$ points. Neither single-factor configuration approaches the pair: the better of the two reaches $90.0\%$ success against the full schedule's $99.0\%$.

Read within each level of the other factor, the two schedules are not interchangeable. The reward schedule is worth $35.0$ points of success when the distance schedule is off and $27.0$ when it is on; the distance schedule is worth $17.0$ points with the reward schedule off and $9.0$ with it on. The reward schedule is thus the larger contributor in both, by a factor of two to three. Its effect also falls disproportionately on success rather than on completion---$+35.0$ against $+3.09$ points with the distance schedule off---so what it supplies is not more of the route traveled but the behavior needed to finish it. That is what the two mechanisms are designed to do respectively: expanding the sampled distance lets the policy reach full route length in graduated steps rather than face it from the first episode, whereas advancing the reward weights tightens the behavior required to complete it once the policy is there---the coordinated scheduling argued for in Section~\ref{subsec:curriculum}.

Table~\ref{tab:curriculum_ablation} reports the end of training only, and an end-point cannot distinguish a configuration that converged to a low score from one that was still improving when the budget ran out. One thing can be said about the no-curriculum control without appealing to its learning curve: it is not short of practice at the mission it is judged on. With the distance schedule disabled it trains at the Stage-4 setting for all $500{,}000$ steps, whereas the full schedule reaches that setting only for its last $150{,}000$ (Section~\ref{subsec:curriculum}), so the control's lower score is not explained by less exposure to the evaluated task. Whether it had also stopped improving is not established here.

\subsection{Ablation: The LiDAR Observation}
\label{subsec:exp_lidar}

The polar histogram raises two questions that the rest of the paper takes for granted: whether the LiDAR stream contributes at all, and how finely it has to be resolved. Both lie on one axis, and Table~\ref{tab:lidar_ablation} sweeps it. The histogram divides a $180^\circ$ field of view into equal sectors, so the bin count fixes both the angular resolution and the size of the observation: $5.6^\circ$ per bin and a $67$-dimensional state at 32 bins, $2.8^\circ$ and $99$ dimensions at 64, and $1.4^\circ$ and $163$ dimensions at 128. Removing the stream altogether leaves the $35$ dimensions of telemetry ($4$), route ($26$) and rule signals ($5$), and shrinks the fused feature from $192$ to $128$. The framework was retrained at every setting with all else held fixed---network depth, PPO hyperparameters, reward constants, curriculum schedule and training budget---and each configuration is evaluated on this subsection's own list of $100$ Stage-4 Town05 episodes (Section~\ref{sec:exp_results}).

Among the three resolutions, navigation performance is insensitive to the choice. The table reports $100$, $99$ and $96$ successes out of $100$ episodes for 32, 64 and 128 bins---a fourfold change in angular resolution for a spread of four episodes---and no pair of these rates is distinguishable at this sample size (Fisher's exact test: $p=1.00$ for 32 versus 64 bins, $0.37$ for 64 versus 128, and $0.12$ for 32 versus 128), so the ordering of the three rows is not a ranking. Because each configuration is a single training run, the test bounds the evaluation sampling error only and does not account for seed-to-seed variation in PPO, which can only widen the intervals.

We attribute this insensitivity to the benchmark rather than to the representation. The evaluation environment is static, so every obstacle the policy must avoid is a fixed element of the road scene whose angular extent at the distances that matter is wide and whose bearing changes only with the ego vehicle's own motion. A $5.6^\circ$ sector already resolves structure of that kind, and subdividing it supplies no distinction the task rewards. The picture is expected to change under dynamic traffic, where what must be resolved is the angular extent and closing bearing of another vehicle---often only a few degrees wide at the range where a yielding or lane-change decision has to be made---and where a bin wide enough to merge that vehicle into the background behind it would discard precisely the cue the decision rests on.

The same table answers the prior question that insensitivity raises: if the resolution does not matter, does the stream? Removing the histogram entirely leaves success at $98.0\%$ against $99.0\%$ with it, a difference of one episode in a hundred that the sample cannot resolve (Fisher's exact test, $p=1.00$), with two collisions against one and a marginally lower lateral deviation. We therefore do not claim that the histogram contributes to the results reported in this paper, and the reason is the one that also makes the bin count irrelevant: the environment is static and the reference route runs along clear roadway, so an episode can be completed without ever localizing an obstacle. Establishing what the histogram is worth requires a benchmark on which an episode cannot be completed without it. The same result bounds what the comparison of Section~\ref{sec:exp_results}-A can be credited to: with the histogram removed, success stays at $98.0\%$, so the margin over the baselines is not carried by the LiDAR stream but by the traffic-rule signals, the stage-aware reward, and the difference in state representation itself, which this study does not separate. What the results do establish about the observation is sufficiency at this input dimension---a $99$-dimensional state carries the success rates and the zero-shot transfer reported above with no images, no point-cloud encoder and no BEV rasterization.

Within the three resolutions, one column does move: mean lateral deviation rises from $0.19\,\mathrm{m}$ at 32 and 64 bins to $0.35\,\mathrm{m}$ at 128. A reading consistent with the rest of the table is capacity rather than information---128 bins enlarge the observation by $64$ dimensions without supplying obstacle detail this task appears to need, and the policy is given the same $500{,}000$-step budget in which to learn the larger input---but this remains a hypothesis consistent with the ordering rather than a demonstrated mechanism. We therefore retain 64 bins for the results reported elsewhere in this paper as the middle setting of the range tested, not as a measured optimum: the evidence above is that no optimum is identifiable in a static environment, and the $2.8^\circ$ sectors it preserves are the ones the dynamic setting described above would call for.

\begin{table}[t]
\centering
\caption{Ablation of the LiDAR Observation. The first row removes the polar histogram entirely; the remaining three vary its angular resolution ($180^\circ$ divided by the bin count). Each row is a separate training run, evaluated with its final checkpoint on one list of $100$ Stage-4 Town05 start--goal pairs replayed identically across the four rows.}
\label{tab:lidar_ablation}
\footnotesize
\setlength{\tabcolsep}{5pt}

\begin{tabular}{cccccc}
\toprule
Bins & Ang.\ Res. & Obs.\ Dim. & $P_{\text{succ}}$ (\%) & $P_{\text{coll}}$ (\%) & $\bar{e}_{\text{lat}}$ (m) \\
\midrule
None & --- & 35 & 98.0 & 2.0 & 0.15 \\
\midrule
32  & $5.6^\circ$ & 67  & 100.0 & 0.0 & 0.19 \\
64  & $2.8^\circ$ & 99  & 99.0  & 1.0 & 0.19 \\
128 & $1.4^\circ$ & 163 & 96.0  & 4.0 & 0.35 \\
\bottomrule
\end{tabular}
\end{table}

\section{Conclusion}
\label{sec:conclusion}

This paper presented CORAL, which trains a goal-directed urban driving policy by advancing a five-stage route-distance curriculum and a stage-aware reward together, on a compact $99$-dimensional route-aware observation that needs neither high-dimensional visual input nor expert demonstrations. In CARLA, CORAL outperformed both PPO baselines by a wide margin at the hardest curriculum stage, reaching the goal in all twenty per-stage evaluation episodes against the baselines' $5\%$ and $10\%$. Transferred zero-shot to seven towns it never trained in, the final policy succeeded in $68$--$98\%$ of episodes on routes of the same $100$--$150\,\mathrm{m}$ length---at or above $93\%$ in four of the seven---against $99.0\%$ in the training town, with mean lateral deviation below $0.35\,\mathrm{m}$. The factorial ablation shows that neither schedule alone matches the full configuration; with both disabled, success falls to $55\%$.

Several limitations bound these results. The evaluation environment is static, and a leave-one-out test removing the LiDAR histogram leaves success at $98.0\%$ against $99.0\%$ with it, so this study does not establish that the histogram contributes to the results above: the benchmark can be solved without ever using the obstacle channel. What these results do establish is that a $99$-dimensional state suffices, with no learned perception front-end. That single evaluation setting---static, and with no variation in weather or lighting---also means the safety behavior shown here concerns fixed obstacles rather than interaction with other vehicles, and leaves untested the illumination and weather robustness that motivates a LiDAR observation in the first place, an argument inherited from the literature rather than shown here. Transfer beyond this simulator and sensor model likewise remains to be shown. Traffic-signal compliance is the weakest behavior reported: the policy stops correctly at a growing share of the red lights it meets as the curriculum proceeds, but still runs about a third of them. Every system and configuration reported here is a single training run; in particular, we cannot demonstrate that the no-curriculum control had converged rather than merely run out of budget.

Evaluating in an environment with obstacles on the route and with dynamic traffic is therefore the most immediate direction, since it addresses those limitations at once, followed by making red-light compliance reliable, quantifying the runtime advantage that the fixed low-dimensional observation is expected to confer, and repeating the study across multiple seeds. Richer semantic scene understanding and multimodal language-model-based driving assistance are natural longer-term extensions.

\bibliographystyle{IEEEtran}

\end{document}